\documentclass[runningheads]{llncs}

\usepackage{eccv}

\usepackage{eccvabbrv}

\usepackage{graphicx}
\usepackage{booktabs}
\usepackage{subfiles}
\usepackage{caption}
\usepackage{bbold}
\usepackage{multirow}
\usepackage{amssymb}
\usepackage{pifont}
\usepackage{float}
\usepackage[symbol]{footmisc}

\newcommand{\datasetname}{RIPTSeg }

\usepackage[accsupp]{axessibility}  %

\usepackage{hyperref}

\usepackage{orcidlink}

\begin{document}

\title{Foundation Model or Finetune? Evaluation of few-shot semantic segmentation for river pollution} 

\titlerunning{Evaluation of few-shot semantic segmentation for river pollution}

\author{Marga Don\inst{1}\orcidlink{0009-0001-5435-8935} 
\and Stijn Pinson \inst{2} \and Blanca Guillen Cebrian \inst{2}
\and Yuki M. Asano \inst{1,3} \\
}

\authorrunning{M.~Don et al.}

\institute{University of Amsterdam, Amsterdam, The Netherlands \and
The Ocean Cleanup, Coolsingel 6 Rotterdam, The Netherlands \and 
Currently at Technical University of Nuremberg\\
\email{marga.don@student.uva.nl, y.m.asano@uva.nl, \\ \{stijn.pinson, b.guillencebrian\}@theoceancleanup.com}
}

\maketitle

\begin{abstract}
  Foundation models (FMs) are a popular topic of research in AI. Their ability to generalize to new tasks and datasets without retraining or needing an abundance of data makes them an appealing candidate for applications on specialist datasets. In this work, we compare the performance of FMs to finetuned pre-trained supervised models in the task of semantic segmentation on an entirely new dataset. We see that finetuned models consistently outperform the FMs tested, even in cases were data is scarce. We release the code and dataset for this work \href{https://github.com/TheOceanCleanup/RiverTrashSegmentation}{here}.
\end{abstract}

\section{Introduction}
In recent years, Foundation Models (FMs) have emerged as a popular focus of research in Artificial Intelligence (AI) \cite{devlin2019bert,kirillov2023segment,brown2020language}. Characterized by their ability to easily generalize to new domains and tasks, FMs offer an exciting opportunity for both research and industry.
From an industry perspective, however, FMs are only preferable when they outperform models specifically trained for a given task. Given that real-life data often differs significantly from data used in research, models trained on existing datasets likely do not match the task at hand. Thus, FMs or finetuning an existing model are logical options, though it is not always evident which is the optimal choice. Finetuned models are said to require substantial amounts of high-quality data for training, which is often not easily available in industry contexts. In such cases, FMs could be the solution. In this work, we investigate whether one should use a FM or Finetune.\\

To properly investigate this question, we require a dataset of images that have not been used in the pretraining stages of any FM. As such, we propose the \datasetname dataset, a real-life dataset containing high-quality images of polluted rivers around the world, alongside high-quality segmentation masks identifying the floating patches of trash in these rivers. The images in \datasetname have not been publicly available before.  

Subsequently, we evaluate two segmentation FMs on this dataset: PerSAM \cite{zhang2023personalize}, a variant of the popular Segment Anything (SAM) model \cite{kirillov2023segment}, and SegGPT \cite{wang2023seggpt}, a generalist segmentation model. We compare these models with a YOLOv8 segmentation model \cite{yolov8_ultralytics}, pretrained on COCO \cite{cocodataset} and finetuned on \datasetname. 

We find that a YOLOv8 model finetuned on at least 30 images from \datasetname outperforms all other tested models, and thus is preferable over FMs, even in cases where data is scarce.

We summarize our main contributions as follows:
\begin{itemize}
    \item We introduce the \datasetname dataset, consisting of data that has not been included in any other dataset previously.
    \item We investigate the trade-off between FMs and Finetuning pre-trained models
    \item We explore methods to refine masks predicted by FMs without additional training
\end{itemize}

\section{Related Work}
\subsection{Segmentation}
In computer vision, segmentation is a fundamental task that involves deciding which object each pixel in an image belongs to. Depending on the specific tasks, the object categories differ. Instance segmentation methods, for example, aim to identify specific instances of objects. In semantic segmentation, categorization is more focused on the semantic category an object belongs to. In this work, we focus on semantic segmentation. 

Recently, large-scale vision models for segmentation have been proposed, inspired by advances made in Natural Language Processing using large-scale models \cite{brown2020language, touvron2023llamaopenefficientfoundation}. For example, the Segment Anything (SAM) model \cite{kirillov2023segment}, allowing the user to prompt SAM by defining a point or box denoting the object of interest. Models based on SAM, like Grounded-SAM \cite{ren2024grounded} or PerSAM \cite{zhang2023personalize}, attempt to adapt SAM to automatically generate prompts based on user input. Another example, Painter \cite{wang2023images}, is a large-scale model capable of adapting to many segmentation task given an example input-output pair. 

\subsection{Trash Detection in Water}
Previous works in trash detection have often focused on identifying the classes of individual objects \cite{GNANN2022118902}, or focus on trash washed up on shore \cite{GONCALVES2020111158, BAO2018388} or floating in the ocean \cite{POLITIKOS2023106466}. As far as the authors are aware, this is the first work to compare FMs and Finetuned models in the task of semantic segmentation regarding trash in rivers.

\subsection{Comparing Foundation and Finetuning models}
In previous research, FMs are often compared to other FMs. The fact that most FMs are zero- or few-shot adaptable to unseen datasets often makes them quick and easy to evaluate, whereas other models must first be finetuned or trained from scratch. As a result, the comparison between FMs and Finetuned models is currently understudied. 

In \cite{mai2023opportunitieschallengesfoundationmodels}, the authors compare several FMs to supervised finetuned models in several tasks in Geospatial data. They find that in text-based settings, FMs can outperform task-specific supervised models. However, on tasks involving images or multimodal data, finetuned supervised models have the upper hand. Working with textual data only, \cite{wornow2023shakyfoundationsclinicalfoundation} surveys the performance of FMs on electronic health records, showing that FMs show improved predictive performance compared to non-FMs. However, they note that FMs should also enable other improvements in clinical settings, such as requiring less labeled data to function, which has not yet been appropriately studied.

\section{Dataset}
\label{sec:dataset}
We propose the \datasetname dataset (RIverine Patch Trash Segmentation), for benchmarking segmentation methods on patches of floating trash. The dataset contains 300 high-resolution images (1944x2592) from 6 different locations (50 images per location) with high-quality ground truth segmentation masks for each image. Several train/test splits of different sizes are defined to allow for reproducible training and testing of models. In addition, the dataset contains predefined candidates for prompting; 5 images and 2 masks of floating trash per image for each location. Note that prompt images used by a model are not included in evaluation.

During labeling, careful consideration was taken to label the patches as precisely as possible. In this context, a single floating object is also seen as a patch. Multiple floating objects were counted as one patch when there was no clear separation between them. Patches smaller than 30 pixels were ignored. Floating non-trash objects (i.e. organic material) were labeled as water.

\begin{figure}[H]
    \centering
    \makebox[\textwidth][c]{\includegraphics[width=1.1\textwidth]{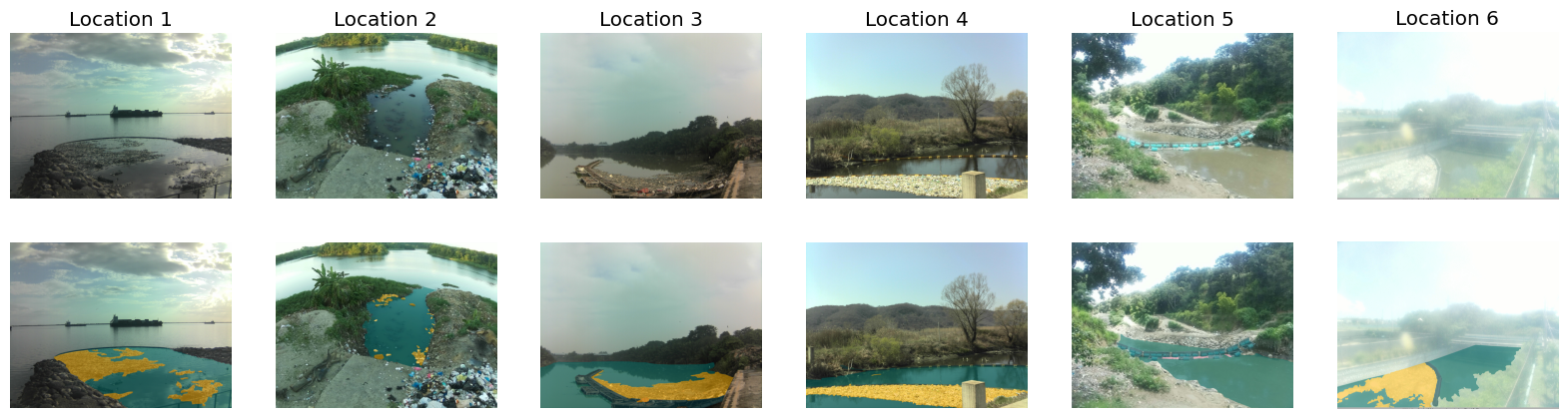}}
    \caption{Example images from the 6 locations in the dataset (upper) with ground truth annotations (lower). Yellow denotes in-system trash, pink denotes out-system trash, light blue denotes water and dark blue denotes the barrier.}
    \label{fig:location-examples}
\end{figure}

In (most of) the rivers included in \datasetname, The Ocean Cleanup has installed barriers called Interceptors to stop the flow of trash downstream. As such, \datasetname contains annotations for 4 classes: water, barrier and trash floating in or out of system. In-system trash is defined as any trash that is floating in the river and upstream of the Interceptor, meaning it can be or has been stopped by the system. In contrast, out-of-system trash is downstream of the barrier.

In total, \datasetname contains 4387 masks. A breakdown of class balance per location can be found in Table \ref{tab:class-distribution}. As indicated in this figure, the 6 locations included in the dataset are quite diverse. Not only does the class balance vary greatly per location, but the sizes of masks vary per class as well. For example, although in-system trash accounts for 52.9\% of all annotations, they only take up 17.4\% of the annotated pixels. We can therefore infer that in-system masks are often relatively small, possibly making them harder to accurately segment. An example image from each location with ground truth segmentation can be found in Figure \ref{fig:location-examples}. Some higher-resolution examples can be found in the supplementary material.

\begin{table}[H]
    \centering
    \scalebox{1}{
    \begin{tabular}{|c|c|c|c|c|c|c|c|c|}
    \hline
    \multirow{2}{4em}{Location ID} & \multicolumn{4}{|c|}{\% mask instances} & \multicolumn{4}{|c|}{\% of pixels in masks} \\
    & In-system & Out-system & Barrier & Water & In-system & Out-system & Barrier & Water \\
    \hline
    Overall & 52.9 & 10.2 & 19.8 & 17.0 & 17.4 & 1.1 & 5.8 & 75.7 \\
    \hline
    1 & 72.4 & 0.0 & 7.3 & 20.2 & 14.4 & 0.0 & 1.0 & 84.6 \\
    2 & 88.5 & 4.4 & 0.0 & 7.1 & 22.0 & 0.9 & 0.0 & 77.1\\
    3 & 62.0 & 18.5 & 8.3 & 11.2 & 14.1 & 2.0 & 17.3 & 66.6 \\
    4 & 21.7 & 3.0 & 53.2 & 22.1 & 16.9 & 0.1 & 5.4 & 77.6 \\
    5 & 12.9 & 25.4 & 44.8 & 16.9 & 2.2 & 4.9 & 10.4 & 82.6\\
    6 & 43.5 & 11.8 & 11.2 & 33.5 & 24.4 & 0.4 & 2.3 & 72.7 \\ 
    \hline
    \end{tabular}
    }
    \caption{Table showing \datasetname statistics. For each location and overall, we present the division of mask instances per class, as well as the proportion of pixels belonging to each class.}
    \label{tab:class-distribution}
\end{table}

\subsection{The Ocean Cleanup}
The data from \datasetname was collected by The Ocean Cleanup, a non-profit organisation aiming to rid the ocean from plastics. Since plastics enter the ocean mostly through rivers \cite{TOC-plastic-ocean}, The Ocean Cleanup has been decreasing the flow of trash into the ocean by cleaning up rivers as well. This is done by installing Interceptors in rivers, which intercept the plastics and other trash on their way downstream. This trash is then extracted and recycled, in order to stay out of the natural environment for good. The images from \datasetname were collected from cameras installed by The Ocean Cleanup to monitor the performance of existing Interceptors, or explore candidate rivers for future deployment. Each image in \datasetname was inspected for depictions of people. If a person was shown on an image, they were digitally removed for privacy purposes. Furthermore, the names of the locations used in \datasetname are not used in this paper or the accompanying codebase.

\section{Methods}

\subsection{RandomForest}
A RandomForest classifier \cite{breiman2001random} is an older technique in the domain of Machine Learning. In short, it trains a set of decision trees, which at test time 'vote' for the most likely class of the given datapoint. In this work, the RandomForest is used as a baseline method, trained on the RGB pixel values of images until convergence. Seeing as the dataset is new, a simple baseline method allows us to compare the more sophisticated models against this baseline. 

\subsection{PerSAM}
PerSAM \cite{zhang2023personalize} is based on the SAM model \cite{kirillov2023segment}, a recent segmentation FM. SAM is used by prompting it with either a point or box prompt, indicating the location of the object(s) to be segmented. However, when attempting to segment objects in many images at once, this would require the user to define a prompt for each image manually. PerSAM attempts to remedy this issue in an efficient, one-shot manner. Given an example image and mask showing the target object, PerSAM finds a location prior for the object using feature similarities. This location prior is then used as a prompt for SAM, resulting in the target object being succesfully segmented. Note that PerSAM uses a frozen SAM model, trained on the extensive SA-1B dataset, making PerSAM itself training-free. 

The original PerSAM \cite{zhang2023personalize} model is designed to predict one mask per prompt\footnote{Note that we use 'prompt' to refer to an image-mask pair.}. However, our data often requires multiple masks to be predicted in a single unseen image. To do so, we add a step to the PerSAM pipeline. PerSAM bases its predictions on feature similarites, captured in the similarity matrix $S \in \mathbb{R}^{h \times w}$, where $h, w$ denote the height and width of the target image. Say we have a predicted mask $M_1 \in \{0,1\}^{h \times w}$. In order to generate another mask, we generate a new similarity matrix $S_1$ as
\begin{equation*}
    S_1 = S \odot (\mathbb{1}_{h,w} - M_1)
\end{equation*}
where $\odot$ denotes the elementwise product and $\mathbb{1}_{h,w}$ a matrix of ones with dimensions $h, w$. This way, we virtually 'black out' the mask $M_1$ from the similarity matrix, meaning PerSAM will likely look to another point in the image to make a new prediction. 

Note that, in theory, we can keep predicting masks this way ad inifinitum. Thus, we need to formulate a condition under which to stop predicting new masks. In this work, we choose to use the mean value of the most recent similarity matrix, the Mean Feature Similarity (MFS). When the MFS falls below a set threshold, this means the image likely does not contain any more candidates for masks and prediction stops. 

In practice, however, we see that PerSAM can still predict the same mask multiple times. Thus, if the MFS barely changes after updating the similarity matrix, we also stop predicting masks, since this indicates that the same mask was predicted twice in a row. 

\subsubsection{Multiple prompts}
In cases where there are multiple patches of different sizes in an image, it could help to use multiple masks to inform the model about the diversity in patch size. Furthermore, multiple prompt images could inform the model about changing conditions, such as patch location or weather. In order to adapt PerSAM to this usecase, we compute the local features for each prompt image and mask. Next, we combine these into one local feature respresentation by taking the mean values over the prompts. The pipeline then continues unchanged.

\subsubsection{PerSAM-F}
An issue with the original SAM is scale ambiguity. Since users prompt SAM using point or box prompts, a user could mean to segment a sub-part of an object instead of the full object. SAM therefore outputs 3 masks of differing scale as options for the user. PerSAM-F, a variant of PerSAM, aims to solve this issue using efficient parameter finetuning. Specifically, the prompt mask is used to finetune 2 weights, which can then adaptively select the correct mask size for future images. In order to use PerSAM-F with multiple prompts per image, we finetune the parameters on multiple prompts instead of only one. Thus, we combine knowledge about the sizes of all prompt masks in the finetuned parameters.

\subsection{SegGPT}
A variant of Painter \cite{wang2023images}, SegGPT is trained specifically for segmentation tasks, as opposed to vision tasks in general. As with Painter, SegGPT is trained using pairs of images and their desired outputs, being segmentation masks in the case of SegGPT. The training procedure involves randomly masking the task output images and training the model to reconstruct the missing pixels. SegGPT was trained on a diverse set of segmentation datasets, including ADE20K \cite{ADE20K}, COCO \cite{cocodataset} and Cityscapes \cite{cordts2016cityscapesdatasetsemanticurban}, to allow generalization to diverse segmentation tasks. 

At inference time, SegGPT is given a prompt image, prompt masks and a target image. Then, SegGPT is able to identify the correct segmentation task based on the prompt mask, and perform this task on the target image. To generate multiple masks for a target image, SegGPT must be prompted repeatedly with a different prompt image or mask. No changes to the pipeline are necessary. Lastly, since SegGPT outputs logits for each prediction, we must threshold the predicted mask to create a binary mask. We refer to this parameter as the Binary Mask Cutoff (BMC). 

\subsection{YOLOv8}
A breakthrough in object recognition, YOLO \cite{redmon2016look} was the first architecture to combine object localization and classification in a single-stage architecture. Since then, many iterations of the model have resulted in YOLOv8 \cite{yolov8_ultralytics}, which is often regarded as the current state of the art in real-time object detection \cite{Solawetz_2024}. The authors of YOLOv8 have also created a segmentation model built on the YOLOv8 architecture, which performs near state-of-the-art on the COCO dataset \cite{cocodataset}.

As a comparison method to the FMs, we finetune a YOLOv8 Segmentation model \cite{yolov8_ultralytics}, pretrained on COCO. Training details can be found in Section \ref{sec:YOLO}.

In order to finetune the model, we combine the in-system and out-system trash classes into one class. However, this means that our model will also segment out-of-system trash. In this case, this is not desirable, since the goal is to estimate only in-system trash. As a remedy, we remove predicted masks based on their location with respect to the barrier. Specifically, we take the predicted masks belonging to the barrier class and compute their mean location. Then, we compute the mean location of each predicted mask and compare it to the location of the barrier. If the mask is located downstream of the barrier, it is removed.

\subsection{Metrics}
\label{sec:metrics-subsection}
The main metric used in this work is mean Intersection over Union (mIoU). 
In most segmentation tasks, IoU is measured for each predicted mask. However, in this case, we are more interested in the IoU of all predicted masks, compared to the ground truth masks per class. Thus, when computing IoU, we combine the predicted masks into one mask and compute the IoU of this mask with respect to the ground truth masks of each class. Our main metric is mIoU-In, the mIoU of the predicted masks with the in-system ground truth masks. We also report mIoU with respect to the other classes, indicating to which degree our model is 'wrong'. A high mIoU-In paired with a high mIoU-Water indicates that the predicted mask contains trash, but also a lot of water, which is undesirable in this task. Thus, mIoU-Water, -Out and -Barrier are better when they are lower.

To gain a further understanding of the performance of models on different  sizes of masks, we divide the ground-truth masks into three categories: small, medium and large, on which we report mIoU-In.

\subsection{Post-hoc mask removal}
In this work, we are only interested in predicting trash in the area of interest, namely the body of water in which patches of trash are found. Predictions outside of this area of interest are not relevant for this task, and therefore not labeled in the dataset. However, models will likely predict masks outside of the area of interest, for example when trash is found on the banks of a river. This will degrade the model's performance, since the IoU of this predicted mask is 0. However, these masks should not influence the metrics, since they are outside of the scope of the task. Thus, we choose to remove all masks predicted outside of the area of interest. Since the cameras used to capture the images never change position, the labeled ground truth masks are used to filter out areas of non-interest. 

\section{Experiments \& results}
\begin{figure}
    \centering
    \makebox[\textwidth][c]{\includegraphics[width=1.15\textwidth]{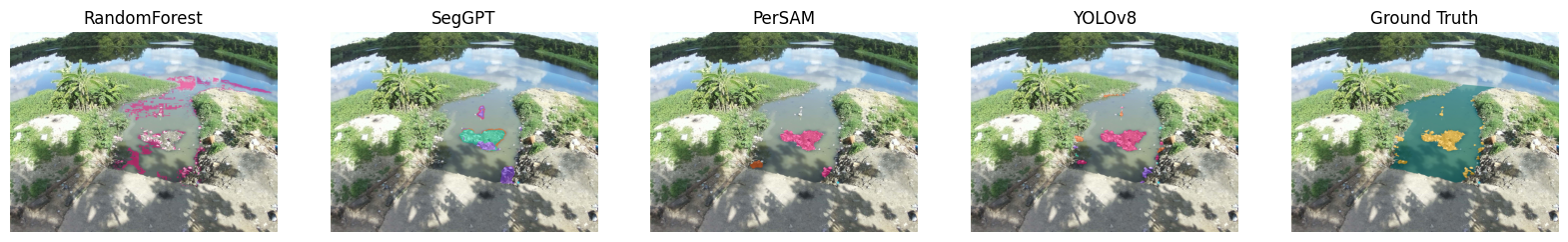}}
    \caption{Example performance of all 4 models using an image from Location 2}
    \label{fig:example_qual}
\end{figure}

In this section, we present the experiments we performed and their results. In-depth discussion of results is reserved for section \ref{sec:discussion}. Table \ref{tab:all-results} shows results of our experiments, detailed below. Example qualitative results are shown in Figure \ref{fig:example_qual}.

From Table \ref{tab:all-results}, we see that the YOLOv8 model consistently outperforms the FMs tested. Note that YOLOv8 nearly doubles performance compared to SegGPT on Location 5. Secondly, SegGPT emerges as the second best model, outperforming PerSAM in most locations. We now discuss experiments performed for each model.
 
\begin{table}[H]
    \centering
    \scalebox{0.9}{
    \begin{tabular}{|c|c|c|c|c|c|c|}
    \hline
    ~ & Location 1 & Location 2 & Location 3 & Location 4 & Location 5 & Location 6 \\ \hline
    \textbf{RandomForest} & ~ & ~ & ~ & ~ & ~ & ~ \\ \hline
    mIoU-In\% & 13.1 & 18.9 & 21.6 & 31.8 & 6.6 & 27.2 \\ \hline
    mIoU-Water\% & 18.7 & 18.6 & 6.2 & 14.9 & 1.0 & 1.2 \\ \hline
    \textbf{SegGPT} & ~ & ~ & ~ & ~ & ~ & ~ \\ \hline
    mIoU-In\% & 46.0 & 45.8 & 60.6 & 72.8 & 24.9 & 73.0 \\ \hline
    mIoU-Water\% & 4.8 & 6.2 & 2.9 & 1.3 & 0.5 & 4.8 \\ \hline
    \textbf{SegGPT + BMC tuning} & ~ & ~ & ~ & ~ & ~ & ~ \\ \hline
    mIoU-In\% & 46.0 & \underline{46.4} & \underline{61.2} & \underline{73.8} & \underline{26.4} & \underline{73.5} \\ \hline
    mIoU-Water\% & 4.8 & 6.2 & 2.6 & 1.1 & 0.5 & 4.4 \\ \hline
    \textbf{PerSAM} & ~ & ~ & ~ & ~ & ~ & ~ \\ \hline
    mIoU-In\% & 16.5 & 29.0 & 25.7 & 42.4 & 2.5 & 24.3 \\ \hline
    mIoU-Water\% & 65.8 & 4.2 & 39.6 & 27.8 & 50.7 & 51.8 \\ \hline
    \textbf{PerSAM-F} & ~ & ~ & ~ & ~ & ~ & ~ \\ \hline
    mIoU-In\% & 49.3 & 23.8 & 39.5 & 65.6 & 6.6 & 31.2 \\ \hline
    mIoU-Water\% & 6.3 & 5.3 & 8.2 & 3.4 & 1.6 & 14.3 \\ \hline
    \textbf{PerSAM-F + MFS tuning} & ~ & ~ & ~ & ~ & ~ & ~ \\ \hline
    mIoU-In\% & \underline{49.3} & 23.8 & 40.3 & 68.6 & 7.3 & 31.2 \\ \hline
    mIoU-Water\% & 1.6 & 5.3 & 7.4 & 5.5 & 4.8 & 14.3 \\ \hline
    \textbf{YOLOv8} & ~ & ~ & ~ & ~ & ~ & ~ \\ \hline
    mIoU-In\% & \textbf{71.7} & \textbf{65.4} & \textbf{77.3} & \textbf{82.9} & \textbf{47.5} & \textbf{87.7} \\ \hline
    mIoU-Water\% & 0.1 & 6.3 & 1.8 & 0.4 & 2.4 & 1.5 \\ 
    \hline
    \end{tabular}}
    \caption{mIoU-In\% and mIoU-Water\% reported for different models and experiment settings. For mIoU-In\% higher is better, for mIoU-Water\% lower is better. All models were evaluated on the pre-defined 40\% test set to allow fair comparison with the YOLOv8 models. Highest mIoU-In\% per location is shown in \textbf{bold}, second best is \underline{underlined}.}
    \label{tab:all-results}
\end{table}

\subsection{RandomForest}
We trained a RandomForest model until convergence for each location using 1-5 training images and corresponding masks as training data. An ablation study identified the highest-performing image combinations, with the best runs reported in Table \ref{tab:all-results}. Both the table and the qualitative analysis in Figure \ref{fig:example_qual} indicate successful model training. However, the masks produced are of poor quality, lacking continuity and semantic coherence, which is to be expected of a RandomForest. Despite this, the model provides a valuable baseline for comparing our other models.

\subsection{SegGPT}
\begin{figure}[H]
    \centering
    \includegraphics[width=0.7\linewidth]{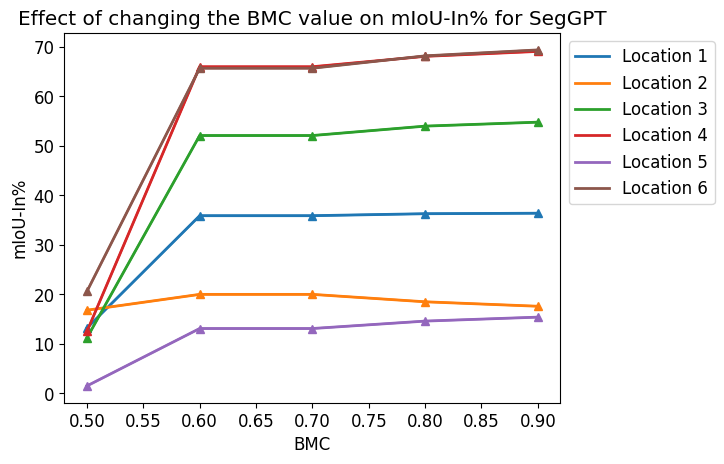}
    \caption{BMC vs mIoU-In\% for SegGPT on all locations, with constant prompt image and mask.}
    \label{fig:SegGPT-BMC}
\end{figure}

Similar to RandomForest, we first identified the most informative prompt for each location using an ablation study with constant BMC. The resulting baseline performance is presented in Table \ref{tab:all-results}, showing improved performance over RandomForest despite disparities between locations. Additionally, Figure \ref{fig:SegGPT-BMC} illustrates the effect of changing the BMC using a constant prompt image, showing that a higher BMC leads to higher mIoU-In\% in most cases.

Using the insights from Tables \ref{tab:all-results} and \ref{fig:SegGPT-BMC}, we aim to find the best combination of prompt image and BMC per location. Shown in Table \ref{tab:all-results} under 'SegGPT+BMC tuning', we see an improvement over the baseline for most locations. 

\subsection{PerSAM}
As with RandomForest and SegGPT, we perform an ablation of the most informative prompt images, for both the training-free version of PerSAM and PerSAM-F. This gives us a baseline performance, shown in Table \ref{tab:all-results} under 'PerSAM' and 'PerSAM-F'. From \ref{tab:all-results}, we see PerSAM-F achieves a higher mIoU-In\% in most locations and a lower mIoU-Water\% in all locations. It must be noted, however, that for Location 5 performance remains low across the board.

\begin{figure}[H]
    \centering
    \includegraphics[width=0.7\linewidth]{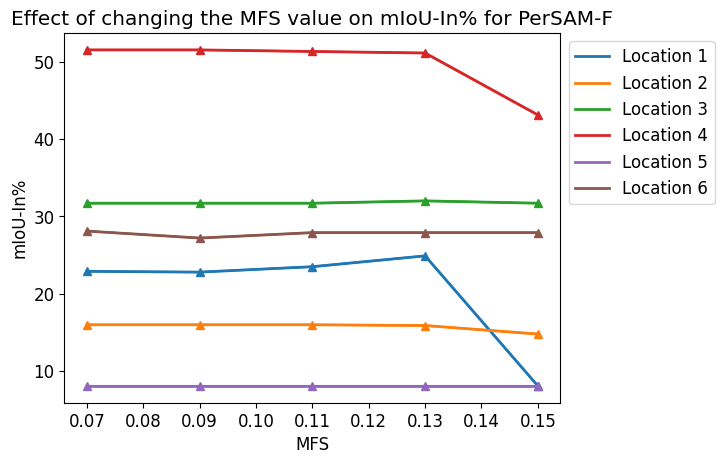}
    \caption{MFS vs mIoU-In\% for PerSAM-F on all locations, with constant prompt image and mask.}
    \label{fig:mmfs}
\end{figure}

To further improve PerSAM-F, we varied the MFS while keeping the prompt image constant. Results are shown in Figure \ref{fig:mmfs}. 

We see that as MFS increases, mIoU-In\% stays relatively stable, until a certain threshold where performance drops for some locations. Note that this is a considerably large drop in performance for only a 0.02 increase in MFS, implying that PerSAM-F is quite sensitive to the specific MFS used. As before, we experiment to find the best combination of MFS and prompt image per location. These results are presented in Table \ref{tab:all-results} under 'PerSAM-F+MFS tuning', showing slight performance gain for most locations.

\subsection{YOLOv8}
\label{sec:YOLO}
As a contrast to the pretrained models, we finetuned a YOLOv8 Segmentation model from Ultralytics \cite{yolov8_ultralytics}. Models were trained for 200 epochs with batch size 4, using the AdamW optimizer \cite{loshchilov2019decoupledweightdecayregularization}, initial learning rate 0.001429, momentum 0.9 and weight decay 5e-4. For the largest training set, 80\% of the dataset, training took 1.8 hours on a single NVIDIA GeForce RTX 3060 Laptop GPU. Note that we trained YOLOv8 on data from all locations. 

\begin{figure}[H]
    \centering
    \includegraphics[width=0.9\textwidth]{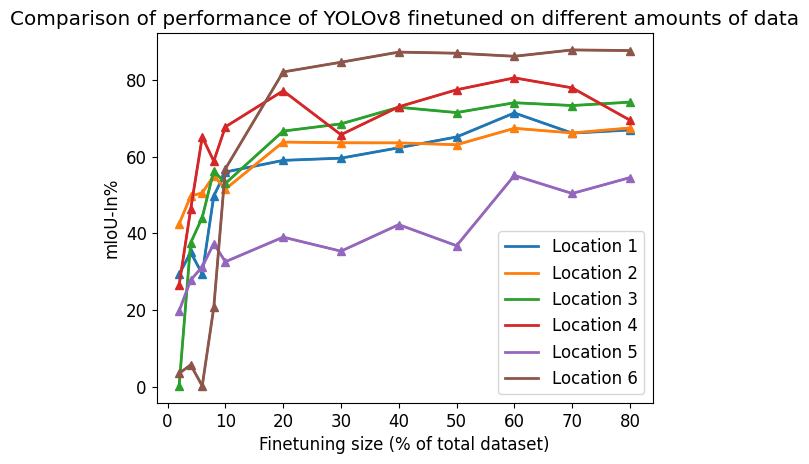}
    \captionof{figure}{Performance of YOLOv8 model finetuned on different sizes of training data. Models were all evaluated on the same 20\% test split.}
    \label{fig:yolo-training-size}
\end{figure}

\newpage
In addition, we trained YOLOv8 on subsets of the training data, while evaluating them on the same test set (20\% of full dataset). Care was taken to ensure that none of the testing datapoints were used in training at any point. These results are shown in Figure \ref{fig:yolo-training-size}. Overall, we see that training on more datapoints increases performance, as is to be expected. However, we also see that for some locations the optimal training set size lies around 60\%, indicating that overfitting could be occurring for larger training sets. Using the model trained with 60\% training data allows us to use 40\% of the dataset as a test set, meaning we can more accurately measure performance on unseen data.

\begin{figure}[H]
    \centering
    \includegraphics[width=0.7\textwidth]{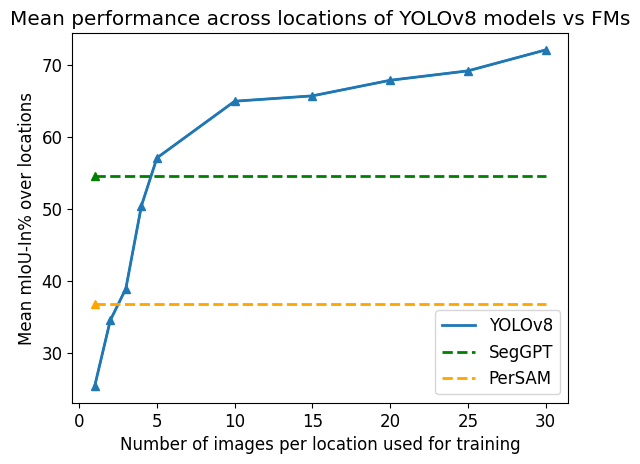}
    \caption{Comparison of YOLOv8 models trained on different training set sizes and the FMs tested. Models were tested on the 40\% test split. For clarity, mIoU-In\% was averaged across locations.}
    \label{fig:yolo-fms-plot}
\end{figure}

In Figure \ref{fig:yolo-fms-plot} we show the performance of YOLOv8 models trained on different training sets compared to SegGPT and PerSAM. We plot mean mIoU-In\% over locations for clarity. We see that YOLOv8 outperforms PerSAM when using 3 images per location, while 5 images are needed to outperform SegGPT. This shows us that even on scarce data, YOLOv8 outperforms the FMs tested.

In Table \ref{tab:all-results}, we compare the YOLOv8 model finetuned on 60\% of the dataset with the other models. Note that for a fair comparison, we evaluate RandomForest, SegGPT and PerSAM at their best settings per location on the 40\% test set used to evaluate the YOLOv8 model. 

\section{Discussion}
\label{sec:discussion}

\subsection{Analysis of Models}
\textbf{RandomForest} performs poorly, as was expected. Masks are of low quality and not semantically coherent. However, qualitative analysis shows RandomForest is often able to correctly identify where patches are located.

\textbf{SegGPT} is the best performing FM we tested. It often produces high-quality, tight masks, but sometimes segments only part of a patch. We believe SegGPT is able to perform strongly on this dataset due to the high similarity between prompt and target images.

\textbf{PerSAM} is promising, but fails in some essential ways. A qualitative analysis shows that PerSAM often predicts  masks consisting fully of water, indicating that the localization of the object of interest is sub-optimal. Note that PerSAM was not originally designed to predict multiple masks, and has shown to be extremely sensitive to hyperparameters controlling this ability. 

\textbf{YOLOv8} is the most succesful of the models tested in this work. It predicts high-quality, tight segmentation masks around patches of trash. However, care must be taken to counter overfitting when finetuning the model. We see that with only 30 training images, YOLOv8 can outperform both SegGPT and PerSAM.

\subsection{Locations}
Throughout all experiments, we see extremely varying performance between locations. At the same time, models seem to agree on which locations are difficult and which are easy: Location 5 is consistently the worst-performing location, while Locations 4 and 6 are often the best. This indicates that the particulars of each location are extremely important. 

An analysis of the locations can give an indication: Locations 4 and 6 view the barrier from upstream, meaning the trash patches aggregate 'in front of' the barrier. Furthermore, the camera is placed quite close to the barrier, meaning the view of the patch is unobstructed and large patches take up a large portion of the image. In contrast, at Location 5 the camera views the barrier from downstream and further away. Thus, part of the patch is obstructed by the barrier and even large patches take up only a relatively small part of the image. In addition, the background in Location 5 often contains washed up trash or rocks which are of a very similar color to the trash, making it difficult to distinguish between patches of trash and background even for a human observer.  

\section{Conclusion}
In this work, we compared the performance of FMs and a finetuned model on a novel dataset. We find that the finetuned model outperforms the FMs, even when finetuning with a limited dataset. In our testing, SegGPT shows impressive generalization capabilities, while PerSAM is not as effective. It is non-trivial to extend PerSAM to predict multiple masks correctly, as it becomes highly sensitive to hyperparameters controlling this ability. 

\subsubsection{Application}
Although \datasetname is a highly specialized dataset, we believe it to be a fair example of a real-life application of segmentation techniques. We show that in such a case, finetuning a model leads to higher performance than using a few-shot FM. Given how small YOLOv8 is compared to the FMs tested and how little data was required to finetune it, finetuning a model is an obvious choice for any real-life dataset.\\ 
It seems that, although the FMs tested show impressive generalization capabilities outside of this work, they are unable to properly adapt to a specialized dataset in a few-shot setting, while real-life datasets are often quite specialized. Thus, further research is needed into the practical applications of FMs.

\subsubsection{Further research}
We encourage researchers to evaluate more models, both FMs and other models, on the \datasetname dataset and improve upon our results. This would further the knowledge of practical applications of FMs, and contribute to cleaner rivers and oceans through improving the quality of data gathered by The Ocean Cleanup.

Aside from evaluating other models, more work can be done to improve the masks from SegGPT. For example, predictions can be refined by prompting SegGPT an additional time with a zoomed-in version of the area of an image where a mask was predicted. This could allow the model to capture masks in more detail. Further exploration of suitable prompts for each image could increase performance as well. Lastly, the possibility of finetuning an FM could be explored, thereby possibly combining the best of both worlds.

\bibliographystyle{splncs04}
\bibliography{main}

\newpage
\section{Supplementary}

\subsection{Ablations}
\subsubsection{Dynamic Prompting}
Firstly, we attempted to work with a 'dynamic prompt' for SegGPT, where each testing image was matched with the closest-matching prompt image. Specifically, we embedded the test and prompt images using a Vision Transformer (ViT) model \cite{dosovitskiy2021imageworth16x16words} and computed the cosine similarity between them. The prompt image with the highest cosine similarity to the test image was chosen as its prompt. We show results in Table \ref{tab:match-finetune-img}. We see that for all locations, mIoU-In decreases when we attempt to match the prompt image to the test image.

\begin{table}[H]
\centering
\begin{tabular}{|c|c|c|}
    \hline
    \multirow{2}{4em}{Location} & \multicolumn{2}{|c|}{mIoU-In} \\
    &  No matching & Matching \\
    \hline
    1 & 46.0 & 45.2 \\
    2 & 46.4 & 41.3 \\
    3 & 61.2 & 52.8 \\
    4 & 73.8 & 69.8 \\ 
    5 & 26.4 & 18.7 \\
    6 & 73.5 & 62.5 \\
    \hline
\end{tabular}
\caption{Results of matching the test image to the closest prompt image using cosine similarity, for SegGPT. For brevity, we report only mIoU-In\%.}
\label{tab:match-finetune-img}
\end{table}

\subsubsection{Patches}
\begin{figure}[H]
    \centering
    \includegraphics[width=\textwidth]{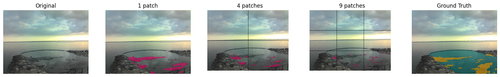}
    \caption{Example showing the effect of dividing images into patches, shown on Location 1.}
    \label{fig:patches}
\end{figure}

Recall that SegGPT compresses the images from 1944x2592 to 448x448, which potentially causes a loss of information. Furthermore, the resulting masks must be upsampled to the original resolution, leading to non-precise segmentations. In an attempt to refine the predictions from SegGPT, we divided each testing image into a grid of patches. This allows us to preserve more image details and leads to more precise segmentation masks. We present quantitative results in Table \ref{tab:seggpt-patches} and qualitative results in Figure \ref{fig:patches}. We see that as the image is divided into more patches, larger patches are being missed. However, SegGPT is able to capture more fine details of smaller patches.

\begin{table}
\centering
\begin{tabular}{|c|c|c|c|}
    \hline
    \multirow{2}{4em}{Location} & \multicolumn{3}{|c|}{\# patches} \\
    & 1 & 4 & 9 \\
    \hline
    1 & \textbf{46.0} & 35.1 & 15.3 \\
    2 & \textbf{45.8} & 38.8 & 29.7 \\
    3 & \textbf{60.6} & 60.3 & 56.7 \\
    4 & 72.8 & \textbf{78.9} & 76.6 \\ 
    5 & \textbf{24.9} & 15.5 & 15.2 \\
    6 & \textbf{73.0} & 68.8 & 57.1 \\
    \hline
\end{tabular}
\caption{Results of dividing the images into a grid of 4 or 9 patches, compared with the best performing run for SegGPT per location as reported in Table \ref{tab:all-results}. For brevity, we report only mIoU-In\%.}
\label{tab:seggpt-patches}
\end{table}

\subsection{Additional metrics}
In Tables \ref{tab:supp_RF}-\ref{tab:supp_YOLO}, we present the models evaluated on the 40\% test split for each location, as in Table \ref{tab:all-results}, with additional metrics. In addition to the metrics described in Section \ref{sec:metrics-subsection}, we report the standard deviation of mIoU per class. Furthermore, we report the mean Hamming distance, which we compute as
\begin{equation*}
    H = \frac{g-p}{g}
\end{equation*}
where $g,p$ denote the number of pixels in the ground truth and predicted masks, respectively. Note that we normalize this with respect to the size of the ground truth mask. This way, the Hamming distance represents the portion of the ground truth mask that is mislabeled. Specifically, we present a positive and negative Hamming distance, representing under- and overestimation of mask size respectively, in order to showcase the different behavior in these cases.

\subsection{Further examples of \datasetname images}
Below, in Figures \ref{fig:example_1}-\ref{fig:example_6} we show further examples from \datasetname, in higher resolution than in section \ref{sec:dataset}.

\newpage
\begin{table}[H]
    \centering
    \scalebox{0.8}{
    \begin{tabular}{|c|c|c|c|c|c|c|c|c|c|}
    \hline
    \multirow{2}{4em}{Location} & \multicolumn{4}{|c|}{mIoU $\pm$ $\sigma$} & \multicolumn{3}{|c|}{Binned mIoU} & \multicolumn{2}{|c|}{Hamming distance} \\
    & In-system & Water & Out-system & Barrier & Small & Medium & Large & Positive & Negative \\
    \hline
    1 & 13.1 $\pm$ 11.2 & 18.7 $\pm$ 9.4 & N.A. & 1.7 $\pm$ 0.9 & 4.4 & 17.1 & 28.8 & 0.3 & 3.8 \\
    2 & 18.9 $\pm$ 13.8 & 18.6 $\pm$ 17.2 & 1.5 $\pm$ 2.8 & N.A. & 7.8 & 18.3 & 37.4 & 0.3 & -4.1 \\
    3 & 21.6 $\pm$ 6.8 & 6.2 $\pm$ 5.0 & 3.6 $\pm$ 3.4 & 16.7 $\pm$ 5.4 & 14.7 & 21.2 & 26.0 & 0.3 & 0.5 \\
    4 & 31.8 $\pm$ 21.6 & 14.9 $\pm$ 18.1 & 0.0 $\pm$ 0.1 & 1.6 $\pm$ 0.9 & 15.8 & 36.4 & 51.1 & 0.3 & 3.0 \\
    5 & 6.6 $\pm$ 7.0 & 1.0 $\pm$ 1.4 & 2.5 $\pm$ 4.2 & 1.5 $\pm$ 1.2 & 0.2 & 8.1 & 7.2 & 0.6 & 6.3 \\
    6 & 27.2 $\pm$ 23.5 & 1.2 $\pm$ 1.3 & 0.4 $\pm$ 0.8 & 0.5 $\pm$ 1.0 & 18.0 & 30.4 & 30.1 & 0.6 & 0.0 \\
    \hline
    \end{tabular}
    }
    \caption{RandomForest evaluated on the 40\% test set for each location. N.A. denotes that a certain class is not present in the ground truth annotations for that location.}
    \label{tab:supp_RF}
\end{table}

\begin{table}[H]
    \centering
    \scalebox{0.8}{
    \begin{tabular}{|c|c|c|c|c|c|c|c|c|c|}
    \hline
    \multirow{2}{4em}{Location} & \multicolumn{4}{|c|}{mIoU\% $\pm$ $\sigma$} & \multicolumn{3}{|c|}{Binned mIoU\%} & \multicolumn{2}{|c|}{Hamming distance} \\
    & In-system & Water & Out-system & Barrier & Small & Medium & Large & Positive & Negative \\
    \hline
    1 & 46.0 $\pm$ 12.4 & 4.8 $\pm$ 2.3 & N.A. & 0.0 $\pm$ 0.0 & 47.8 & 42.5 & 46.6 & 0.3 & 0.5 \\
    2 & 45.8 $\pm$ 15.6 & 6.2 $\pm$ 3.0 & 0.1 $\pm$ 0.4 & N.A. & 41.3 & 53.4 & 42.2 & 0.4 & 0.9 \\
    3 & 61.2 $\pm$ 10.3 & 2.6 $\pm$ 3.3 & 0.0 $\pm$ 0.1 & 7.3 $\pm$ 2.9 & 57.5 & 57.0 & 68.8 & 0.1 & 0.3 \\
    4 & 73.8 $\pm$ 15.2 & 1.1 $\pm$ 0.6 & 0.0 $\pm$ 0.1 & 9.4 $\pm$ 7.3 & 56.6 & 86.2 & 84.1 & 0.0 & 0.3 \\
    5 & 26.4 $\pm$ 17.6 & 0.5 $\pm$ 0.7 & 0.2 $\pm$ 0.8 & 6.0 $\pm$ 3.9 & 2.0 & 31.2 & 29.3 & 0.3 & 4.6 \\
    6 & 73.5 $\pm$ 13.9 & 4.4 $\pm$ 3.9 & 1.5 $\pm$ 3.0 & 0.4 $\pm$ 0.6 & 57.8 & 77.8 & 80.1 & 0.1 & 0.2 \\
    \hline
    \end{tabular}
    }
    \caption{Best runs from SegGPT evaluated on the 40\% test set for each location. N.A. denotes that a certain class is not present in the ground truth annotations for that location.}
    \label{tab:supp_SegGPT}
\end{table}

\begin{table}[H]
    \centering
    \scalebox{0.8}{
    \begin{tabular}{|c|c|c|c|c|c|c|c|c|c|}
    \hline
    \multirow{2}{4em}{Location} & \multicolumn{4}{|c|}{mIoU\% $\pm$ $\sigma$} & \multicolumn{3}{|c|}{Binned mIoU\%} & \multicolumn{2}{|c|}{Hamming distance} \\
    & In-system & Water & Out-system & Barrier & Small & Medium & Large & Positive & Negative \\
    \hline
    1 & 49.3 $\pm$ 21.1 & 1.6 $\pm$ 1.0 & N.A. & 0.0 $\pm$ 0.1 & 57.2 & 40.1 & 43.4 & 0.4 & 0.2 \\
    2 & 23.8 $\pm$ 24.9 & 5.3 $\pm$ 17.2 & 0.0 $\pm$ 0.0 & N.A. & 14.6 & 36.6 & 20.7 & 0.7 & 8.0 \\
    3 & 40.3 $\pm$ 25.8 & 7.4 $\pm$ 13.6 & 0.7 $\pm$ 2.2 & 1.5 $\pm$ 2.9 & 52.9 & 40.5 & 32.7 & 0.5 & 0.8 \\
    4 & 68.6 $\pm$ 29.9 & 5.5 $\pm$ 15.8 & 0.0 $\pm$ 0.0 & 6.2 $\pm$ 10.9 & 49.2 & 73.0 & 93.6 & 0.1 & 2.5 \\
    5 & 7.3 $\pm$ 8.7 & 4.8 $\pm$ 4.9 & 0.1 $\pm$ 0.4 & 0.1 $\pm$ 0.2 & 0.1 & 3.5 & 16.0 & 0.8 & 12.0 \\
    6 & 31.2 $\pm$ 25.5 & 14.3 $\pm$ 15.3 & 0.0 $\pm$ 0.0 & 0.2 $\pm$ 0.5 & 19.1 & 28.2 & 45.8 & 0.7 & 1.2 \\
    \hline
    \end{tabular}
    }
    \caption{Best runs from PerSAM-F evaluated on the 40\% test set for each location.N.A. denotes that a certain class is not present in the ground truth annotations for that location.}
    \label{tab:supp_PerSAM}
\end{table}

\begin{table}[H]
    \centering
    \scalebox{0.8}{
    \begin{tabular}{|c|c|c|c|c|c|c|c|c|c|}
    \hline
    \multirow{2}{4em}{Location} & \multicolumn{4}{|c|}{mIoU\% $\pm$ $\sigma$} & \multicolumn{3}{|c|}{Binned mIoU\%} & \multicolumn{2}{|c|}{Hamming distance} \\
    & In-system & Water & Out-system & Barrier & Small & Medium & Large & Positive & Negative \\
    \hline
    1 & 71.7 $\pm$ 10.1 & 2.4 $\pm$ 1.0 & N.A. & 0.0 $\pm$ 0.1 & 66.0 & 72.9 & 84.2 & 0.1 & 0.2 \\
    2 & 65.4 $\pm$ 19.0 & 6.3 $\pm$ 5.9 & 2.1 $\pm$ 5.0 & N.A. & 50.2 & 74.3 & 77.2 & 0.1 & 0.4 \\
    3 & 77.3 $\pm$ 7.1 & 1.8 $\pm$ 1.4 & 0.4 $\pm$ 1.0 & 1.2 $\pm$ 1.1 & 73.6 & 76.5 & 80.3 & 0.1 & 0.1 \\
    4 & 82.9 $\pm$ 14.3 & 0.4 $\pm$ 0.4 & 0.0 $\pm$ 0.0 & 0.9 $\pm$ 1.0 & 70.6 & 90.3 & 92.3 & 0.1 & 0.1 \\
    5 & 47.5 $\pm$ 29.1 & 0.1 $\pm$ 0.2 & 2.4 $\pm$ 5.9 & 0.6 $\pm$ 0.5 & 19.4 & 62.4 & 38.2 & 0.5 & 0.3 \\
    6 & 87.7 $\pm$ 4.9 & 1.5 $\pm$ 1.3 & 0.0 $\pm$ 0.0 & 0.3 $\pm$ 0.5 & 88.0 & 86.0 & 89.8 & 0.1 & 0.0 \\
    \hline
    \end{tabular}
    }
    \caption{YOLOv8 evaluated on the 40\% test set for each location. N.A. denotes that a certain class is not present in the ground truth annotations for that location.}
    \label{tab:supp_YOLO}
\end{table}

\begin{figure}[H]
    \centering
    \includegraphics[width=0.9\textwidth]{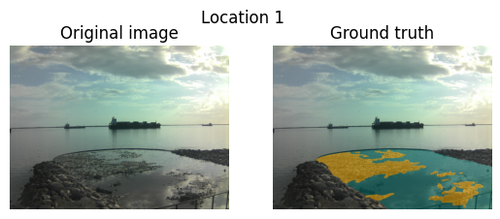}
    \caption{Example image of Location 1.}
    \label{fig:example_1}
\end{figure}
\begin{figure}[H]
    \centering
    \includegraphics[width=0.9\textwidth]{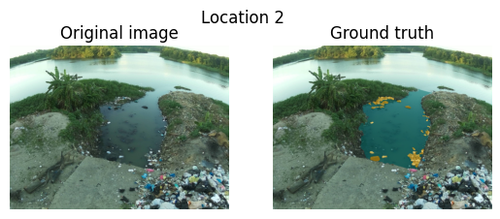}
    \caption{Example image of Location 2.}
    \label{fig:example_2}
\end{figure}
\begin{figure}[H]
    \centering
    \includegraphics[width=0.9\textwidth]{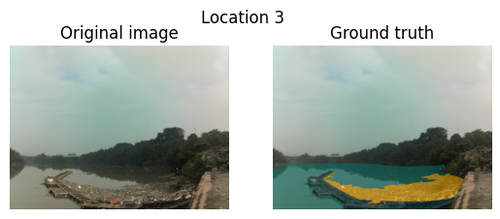}
    \caption{Example image of Location 3.}
    \label{fig:example_3}
\end{figure}
\begin{figure}[H]
    \centering
    \includegraphics[width=0.9\textwidth]{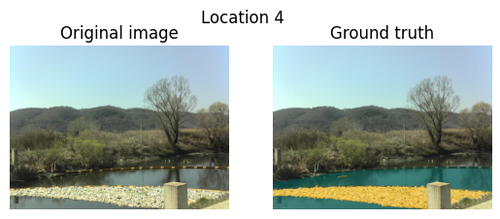}
    \caption{Example image of Location 4.}
    \label{fig:example_4}
\end{figure}
\begin{figure}[H]
    \centering
    \includegraphics[width=0.9\textwidth]{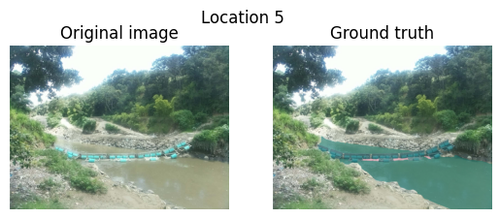}
    \caption{Example image of Location 5.}
    \label{fig:example_5}
\end{figure}
\begin{figure}[H]
    \centering
    \includegraphics[width=0.9\textwidth]{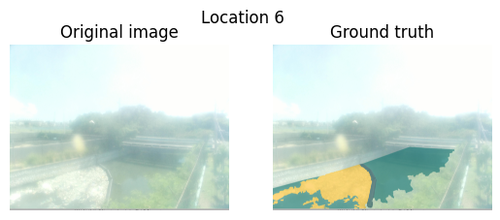}
    \caption{Example image of Location 6.}
    \label{fig:example_6}
\end{figure}

\end{document}